\documentclass[letterpaper, 10 pt, conference]{ieeeconf}  
\usepackage{blindtext}
\usepackage{graphicx}

\usepackage{caption}
\usepackage{subfig}
\usepackage{amsmath} 
\usepackage{amssymb}  
\usepackage{bm} 
\usepackage{tikz}
\usepackage{verbatim}
\usepackage{epstopdf}
\usepackage{tabularx} 
\usepackage{url}

\IEEEoverridecommandlockouts                              

\title{\LARGE \bf
Multi-session Map Construction in Outdoor Dynamic Environment*
}

\author{Xiaqing Ding$^{1}$, 
Yue Wang$^{1}$, 
Huan Yin$^{1}$,
Li Tang$^{1}$,
Rong Xiong$^{1}$
\thanks{*This work was supported in part by the National Nature Science Foundation of China under Grant U1609210, in part by the Science Fund for Creative Research Groups of NSFC under Grant 61621002, and in part by the Man-Machine Cooperative Mobile Dual-Arm Robot for Dexterous Operation in Intelligent robot special of national key research and development program under Grant 2017YFB1300400. }
\thanks{$^{1}$Xiaqing Ding, Yue Wang, Huan Yin, Li Tang, Rong Xiong are with the State Key Laboratory of Industrial Control and Technology, Zhejiang University, Hangzhou, P.R. China. Yue Wang is with iPlus Robotics Hangzhou, P.R. China. Yue Wang is the corresponding author {\tt\small wangyue@iipc.zju.edu.cn}. Rong Xiong is the co-corresponding author {\tt\small rxiong@zju.edu.cn}.}%
}

\begin{document}

\maketitle
\thispagestyle{empty}
\pagestyle{empty}

\begin{abstract}
Map construction in large scale outdoor environment is of importance for robots to robustly  fulfill their tasks. Massive sessions of data should be merged to distinguish low dynamics in the map, which otherwise might debase the performance of localization and navigation algorithms. In this paper we propose a method for multi-session map construction in large scale outdoor environment using 3D LiDAR. To efficiently align the maps from different sessions, a laser-based loop closure detection method is integrated and  the sequential information within the submaps is utilized for higher robustness. Furthermore, a dynamic detection method is proposed to detect dynamics in the overlapping areas among sessions of maps. We test the method in the real-world environment with a VLP-16 Velodyne LiDAR and the experimental results prove the validity and robustness of the proposed method. 

\end{abstract}

\section{INTRODUCTION}

Substantial progresses in SLAM enable robots to construct an accurate and even a dynamic-free map within a single session using 3D LiDAR in outdoor environment \cite{pomerleau2014long}. When the environment is large, the map is usually represented as a conjunction of linked submaps, in which way the computation burden could be restricted under a sustainable range, which benefits the performance of SLAM and localization algorithms. However, a single-session map is not adequate for long-term operation tasks, especially in outdoor environment where the dynamics can not be suppressed. Besides the moving objects, there are many low-dynamic objects in outdoor environment, such as the parking cars, which keep still throughout a whole  session but are potential to change positions across sessions. If no priori information is provided, they can not be distinguished within a single session and are remained in the constructed map. Those low dynamic objects in the map sometimes would introduce wrong data association results for localization algorithms as some  part of the observation can not find correct correspondences in the map. What's more, they also deteriorate the performance of navigation algorithms as some free space is explicitly occupied in the map. To improve the performance of outdoor tasks, multi-session of data is required to remove the dynamics in the map.

As for multi-session map maintenance, one of the most important issues is how to efficiently align maps across sessions. Maps in different sessions need to be conjuncted into a whole map after alignment and a global optimization could be applied to improve the consistency of the map. This can also be regarded as a broader loop closure detection problem, while in terms of laser-based algorithms, it still remains to be a difficult problem due to the high ambiguity of 3D point clouds.

\begin{figure}[thpb]
\centering
\includegraphics[width=0.5\textwidth]{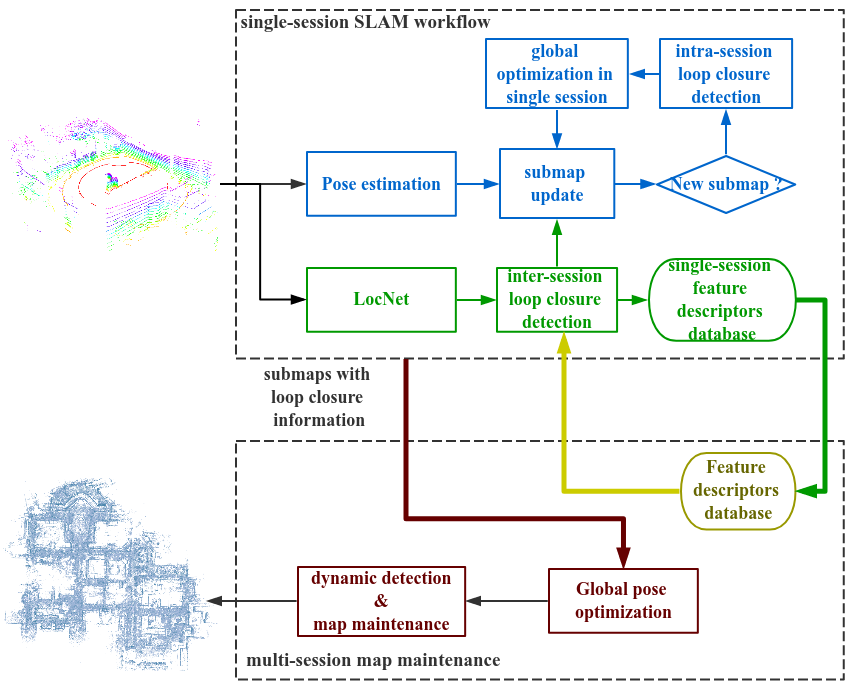}
\caption{The framework of the multi-session laser-based map maintenance method. }
\label{framework}
\end{figure}

In this paper, a multi-session map maintenance method is developed in large scale outdoor environment using the 3D LiDAR. The map is conjuncted from linked submaps and the intra-session loop closure detection is performed based on the poses of submaps following an overlap-based outlier detection. Extended from our previous work, information from the history sessions is converted into feature vectors and organized into a database for quick index.
The inter-session loop closure is detected for each scan based on the database  while performing laser-based SLAM method in a new session and sequence information is used in order to improve the robustness. What's more, a new dynamic detection method designed for detecting in sparse point clouds is proposed to remove the dynamics out of the map. The main contributions of this paper are as follows:

\begin{itemize}
\item  Introduce a multi-session laser-based map construction framework which is designed for serving localization and navigation tasks in large-scale outdoor environment.
\item Integrate the laser-based loop closure detection method in our previous work into the multi-session map construction framework for map alignment based on submaps and introduce sequential information to improve the robustness.
\item Propose a low-dynamic detection method between sparse point clouds from different sessions to maintain only the static parts in the map, which aims at suppressing the influence from the dynamics in the environment while localizing and navigating against the priori map.
\end{itemize}

The remain of the paper is organized as follows: Section II gives a review of the related works and the followed Section III introduces the system overview of the proposed multi-session map construction method. Section IV, V, VI separately demonstrate the single-session laser-based SLAM method, inter-session loop closure detection method and the multi-session map construction method implemented in the proposed system. And in Section VII the experimental results are presented. Conclusion and some discussions are showed in Section VIII.

\section{Related Work}
To bring the SLAM algorithms into practice, many researches focus on how to improve the robustness in terms of long term operation. Two of the relevant issues are addressed in this paper: one is how to efficiently align the multi-session results into a consistent map and the other is how to extract the stable parts of the environment according to  massive sessions of data.  

To realize multi-session map alignment, robots should be capable of detecting loop closure globally and across sessions. With the development of computer vision and machine learning, vision-based loop detection method is gradually mature in recent years. McDonald et al. \cite{mcdonald20116} introduce anchor nodes to combine the pose graphs from massive sessions based on the loop closures detected from the visual-based place recognition method. Recently Schneider et al \cite{schneider2017maplab} propose an impressive visual-inertial mapping and localization framework named maplab. They use an efficient binary description for loop detection then add the information as edges into the posegraph for optimization.  While in context of laser-based methods, there still exists large difficulties. Yang et al. \cite{yang2013go} propose a global point cloud registration method expanding from ICP algorithm \cite{besl1992method} with Branch and Bound \cite{land1960automatic} scheme. Though efficiently, it's still hard to be applied in large scale outdoor environment. SegMatch \cite{dube2017segmatch} extracts segments from 3D point cloud and tries to match them with segments extracted from history trajectories for loop candidate detection. Being different from SegMatch that utilizes features in the environment for detection, Yin et al. \cite{yin2017efficient} introduce LocNet to describe the whole range information of a 3D point cloud for matching, which reveals good performance in unstructured outdoor environment. In this paper, we also introduce  LocNet for point cloud representation and detect loop closure for each laser scan against the  history point cloud information indexed in the database.

In the context of long-term localization and navigation, it's essential that robots should be capable of distinguishing dynamics out of the static environment. Walcott-Bryant et al. \cite{walcott2012dynamic} propose the 2D dynamic pose graph SLAM and detect dynamics across time on the occupancy grid map. 
While in 3D space, Fehr et al. \cite{fehr2017tsdf} represent the environment using tsdf-based structure, which could inherently distinguish the empty voxel out of the unobserved one. Pomerleau et al. \cite{pomerleau2014long} update the dynamic probabilities of the point cloud based on the observations from each new scan while incrementally mapping, they can also to large extend suppress dynamic points from adding into the maps. In this paper we try to detect the dynamics between two sparse submaps in outdoor environment. Instead of directly converting the whole submap into a tsdf-based map, we propose a new detection method which could largely decrease the computation burden of building the voxel map and designed especially for dynamic detection in multi-view merged submaps.

\section{System Overview}
The overview of the large-scale multi-session mapping framework will be introduced in this section. The framework includes a single-session SLAM workflow, a multi-session map maintenance system and a loop closure detection pipeline for cross-session loop closure detection among submaps, as shown in Fig. \ref{framework}. 

In consideration of real-time performance, while performing the laser-based SLAM algorithm, the whole map is organized by linked submaps, which are accumulated by sequential laser scans following the map updating method in \cite{pomerleau2014long} and little high-dynamic points would be remained in the submaps. When a new submap is established, the intra-session loop closure detection module will search the loop closure candidates across the history submaps within this session based on the priori pose estimations, following a overlapping-based loop closure validation step. If a loop closure is detected and validated, a global pose optimization will be applied to improve the accuracy and consistency of the SLAM results.

Besides loop closure detection within a single session, we introduce the LocNet \cite{yin2017efficient} for cross-session loop closure detection. LocNet converts the range information of each scan into a compact feature vector and all of the feature vectors will be maintained in an indexed database which enables quick matching among large numbers of vectors. Therefore when performing a new session of SLAM algorithm, a quick place recognition process can be implemented for each scan based on the existing database constructed from the information of massive history data and the sequential detection information within each submap is utilized for the inter-session loop closure validation. 

When a new session of map is constructed, we introduce the inter-session loop closures as edges into the global pose optimization framework to align the new session of map into the history one. 
After the map alignment, dynamics will be detected within the overlapping areas between each pair of new and history submaps using our proposed sparse point cloud based dynamic detection method, which remains a more static map that benefits future localization and navigation tasks.

\section{Single Session Laser-based SLAM}
3D LiDAR could provide precise geometry information from the environment, which makes the pose estimation more accurate compared with the vision-based method. As for outdoor environment, there exists many high dynamic objects such as the moving cars and pedestrian, which should not be included in the constructed map. In this section we utilize the ICP method implemented from \cite{Pomerleau12comp} for pose estimation and follow the map updating method introduced in \cite{pomerleau2014long} to accumulate the map.

In consideration of real-time performance when the exploration area is large, we only perform the SLAM algorithm on a submap and record the relationship between the neighbor submaps for the global pose graph optimization. When the relative transformation between the current pose and the pose of current submap origin is beyond the designed threshold, the current submap is regarded as finished and a new submap is initialized on the pose of the current frame.

When a submap is finished, the intra-session loop closure detection method will be executed. Some candidates might be selected based on the priori pose estimation information of the existing submap origins. If two submap origins are close in Euclidean space, they are with high probability to claim loop closure and are regarded as a candidate pair. We then apply the ICP algorithm on the two submaps to compute their relative transformation. If the result approximates their priori relative transformation derived from the priori pose estimations and the overlap ratio between the two submaps is reasonable after the iteration result of ICP algorithm, this candidate pair passes the validation process and is regarded as a loop closure. Then a global pose graph optimization is performed

 \begin{equation}
 \begin{split}
&\left\{x_i|x_i\in{X}\right\} = \\
\arg\min_{x_i} &\sum\limits_{x_i} \sum \limits_{x_j \in{N_{x_i}}}{\rho(||(x_i\ominus x_j) \ominus z_{i,j}||^2_{\Omega_{i,j}})}
 \end{split}
 \end{equation}
 where $X=\left\{x_0, x_1...x_n\right\}$ represents the pose estimation set of submap origins within current session, $N_{x_i}$ demonstrates the neighbor set of submap $x_i$, which includes both the successive submaps while mapping and the submaps introduced by the loop closures. The notation $x_i\ominus x_j$ means the relative transformation between pose $x_i$ and pose $x_j$ as demonstrated in \cite{lu1997globally}.  $z_{i,j}$ represents the observation results computed from the ICP algorithm between the successive or loop closed submaps with the pose of $x_i$ and $x_j$. $\rho(\cdot)$ is the robust kernel and $\Omega_{i,j}$ represents the information matrix. The whole optimization function is solved using 
 Levenberg-Marquardt algorithm \cite{K2011G2o}.

\section{inter-session Loop closure detection}
In our previous work \cite{yin2017efficient}, a global localization method is proposed which converts the range information of each laser scan into a feature vector using LocNet then place recognition is achieved by matching the current feature vector with the indexed database that contains the information of the whole map. In this paper we integrate this method into multi-session map maintenance framework for inter-session loop closure detection.

Initially, in the first session we convert each laser scan into the feature vector while performing the SLAM algorithm. In the end besides saving the submaps and their relationships for map maintenance, we also save all of the feature vectors and index them into a loop closure database using the kd-tree algorithm. When a new session of SLAM algorithm is performed, after converting the current laser scan $s_i^c$ into the feature vector $l_i^c$, we search the closest feature vector $l_j^h$ in the kd-tree based database, which indicates the most resemble laser scan $s_j^h$ in history sessions. We record the submap $m_k^h$ which the laser scan $s_j^h$ belongs to as a label of the current laser scan $s_i^c$. If the distance between the matched vectors is larger than $d_\alpha$, we consider there is no matching history information for the current scan.  

When this session of data has been processed, we select the inter-session loop closure candidates by statistically checking the place recognition results within each new submap in a voting way based on the following criterion
\begin{itemize}
\item Sum up all of the submaps that are matched by the laser scans in current submap $m^c_i$. Compute the proportions $p_j$ for each matched submap $m_j^h$.
 \begin{equation}
 \begin{split}
p_j = \frac{number \  of \ scans \ that \ match \ with  \ m_j^h}{number \ of \ the \ scans \ in \ m^c_i}
 \end{split}
 \end{equation}
\item If the largest proportion $p_j$ is larger than $\gamma$, the current submap $m_i^c$ and the history submap $m_j^h$ are considered as  a pair of loop closure candidate.
\item If the sum of top $n$ largest proportions  is larger than $\gamma$, and the corresponding matched submaps are neighbors, all of the matched submaps are considered as the loop closure candidates to the current submap $m^c_i$.
\end{itemize}

After selecting the loop closure candidates, the global ICP algorithm \cite{yang2013go} is applied to compute the relative transformation between the two submap within each candidate pair. The overlap-based outlier detection is also utilized to remove the wrong loop closures as mentioned in section IV.

\begin{figure*}[!htpb]
\centering
\includegraphics[width=0.99\textwidth]{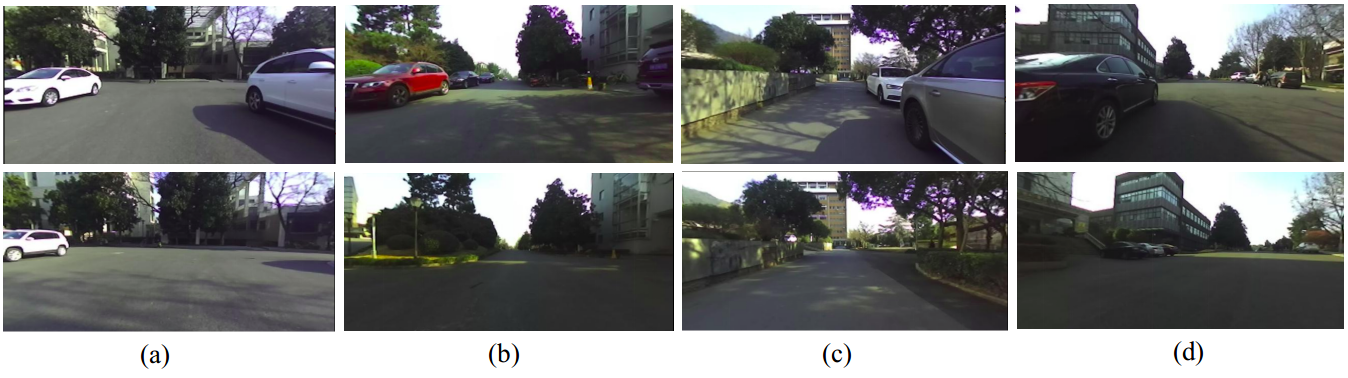}\hfill
\caption{\label{near dynamic} Four places where the inter-session loop closure detection method fails. The pictures in the first line are captured from the  sequence Day 3-1. And the pictures in the second line are the corresponding places in sequence Day 1-1.}
\end{figure*}

\section{Multi-session laser-based SLAM}
This section will introduce the multi-session map maintenance method in our framework. When a new session of map has been constructed and the inter-session loop closures have been detected and validated, we utilize the loop closures as indications to achieve map alignment. Further the dynamics in the overlapping area between the matched new and history submaps will be detected and removed.

\subsection{Multi-session global optimization}
We add the loop closure information as edge into the global pose graph. Mathematically, we denote  the pose set of the history submap origins as $X^h$, the pose set of the submap origins in the new session as $X^c$, the set of inter-session loop closures as $O=\left\{o(x^c_i,x^h_j)\right\}_{x^c_i \in{X^c}, x^h_j \in{X^h}}$.  The optimization function is 
\begin{equation}
\begin{split}
&\left\{x_i|x_i\in{(X^c\cup X^h)} \right\} = \\
\arg\min_{x_i} (&\sum\limits_{x_i} \sum \limits_{x_j \in{N_{x_i}}}{\rho(||(x_i\ominus x_j) \ominus z_{i,j}||^2_{\Omega_{i,j}})} + \\
&\sum \limits_{o(x^c_p,x^h_q)\in{O}}{\rho(||(x^c_p\ominus x^h_q) \ominus z^o_{i,j}||^2_{\Omega^o_{i,j}})})
\end{split}
\end{equation}

where the $z^o_{i,j}$ means the relative transformation computed using the ICP algorithm between the loop closed submaps. We only set the origin of the first submap as the origin of the whole map, therefore the whole trajectory of the new session will be aligned into the same coordination of the history map.

After the alignment of the new session, we once again check loop closure for each submap in the new session as mentioned in IV to expand the set $O$. And the global pose graph optimization would further be applied for higher accuracy and consistency.

\subsection{Dynamic detection in sparse point clouds}
The laser point clouds in the submaps are relatively sparse than  the point clouds constructed with RGB-D sensor \cite{fehr2017tsdf,KahlerPM16}, and the scale is also much larger than the general RGB-D map. So we can not directly apply the existing dynamic detection method in our framework. In this section we introduce a voxel-based dynamic detection method which is designed to process the sparse point clouds.

For each pair of loop closed submaps $(m^h_i,m^c_j)$ indicated in set $O$,  we consider every submap is constructed from two subsets which we denote as $m^h_i = \left\{m^{hc}_i, \overline{m^{hc}_i} \right\}$,  $m^c_j = \left\{m^{hc}_j, \overline{m^{hc}_j} \right\}$. Insides $m^{hc}_i$ and $m^{hc}_j$ represent the common parts in both of the submaps; while $\overline{m^{hc}_i}$ and $\overline{m^{hc}_j}$ indicate the different parts of the two submaps. The dynamic detection process includes the following steps
\begin{itemize}
\item Transform the submap $m^h_j$ into the local frame of submap $m^c_i$.
\item  Construct a kd-tree from the submap $m^h_j$ to search the closest point in the submap $m^c_i$ for each point in $m^h_j$. Keep those matched point pairs that their distances are shorter than $d_\beta$ as seeds.  
Further a region-growing method is applied separately in $m^c_i$ and $m^h_j$ based on the two sets of seeds to determine the subsets ${m^{hc}_i}$ and $m^{hc}_j$, while during growing both the distance between points as well as the difference of normal vectors are considered. Also $ \overline{m^{hc}_i}$ and $\overline{m^{hc}_j}$ are determined as the complementary sets of ${m^{hc}_i}$ and $m^{hc}_j$.
\item Build a voxel map based on the subset $\overline{m^{hc}_j}$. Use the points in $\overline{m^{hc}_j}$ for raycasting and only keep those occupied voxels which are in the front of each raycasting line. Apply the same process on $\overline{m^{hc}_i}$.   
\item Use the remained occupied voxels building from $\overline{m^{hc}_i}$ for raycasting and coarse dynamic points could be identified.  Then apply a region-growing method to get the final detected dynamic points. Record the points in $\overline{m^{hc}_i}$ that indicate the dynamics as $\widehat{m^{hc}_i}$.
\end{itemize}

To update the submap $m^h_j$, remove the detected dynamics  and add the points in $\widehat{m^{hc}_i}$ which indicate the dynamics into it for mending the points behind the dynamics.  Keep the remaining submaps in the new session that do not share large overlapping with the hisory map and update the whole relative relationship among the submaps.

\section{Experimental Results}

\begin{figure*}[!htpb]
\centering
\includegraphics[width=0.99\textwidth]{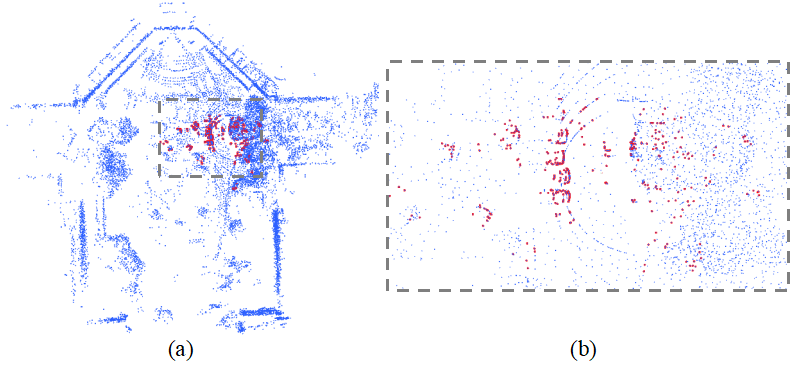}
\caption{The dynamic detection result in a submap. The blue points represent the origin laser submap. The red points represent the detected dynamic points. (b) is the enlarged figure of the dotted area in (a).}
\label{low dynamic detection}
\end{figure*}

In this section we test our multi-session map construction and maintenance method in real-world environment. The data is collected with a four-wheel mobile robot equipped with a VLP-16 Velodyne LiDAR. All of the algorithms are employed in an laptop with  Intel i7-6700HQ @ 2.60GHz * 8 CPU and 7.7 GB memory. The environment we aimed for constructing the map is the south part of Yuquan Campus, Zhejiang University, China, which occupies around 250000 square meters. 

\subsection{Inter-session loop closure detection}
We first test the performance of inter-session loop closure method.  For comparison with the single frame loop closure detection results in \cite{yin2017efficient}, we also test the submap-based loop closure detection on two sessions of Day 3 in YQ21 dataset. YQ21 dataset is a 21-session dataset recorded in March 3, 7 and 9 in 2017, and each session follows the same routine which is around 1.1km. While performing the single-session SLAM algorithm, we set the translation threshold to create a new submap as 20 meters and rotation threshold as 0.7 rad.

The loop closure detection results of every frame are recorded and the performance is demonstrated in \cite{yin2017efficient}. We set the threshold $\gamma$ as 0.95 and $n$ as 3. It's important to note that in some frames where there are many dynamic objects moving around the robot, the loop closure detection method might fail to converge since the range information is largely influenced. Also when detect loop closure across sessions to merge a larger map, the distance between the current feature vector collected from a never visited place and the searched feature vector in the database is relatively large. In this case the the loop closure detection method also returns a failure signal. And for those frames the results are excluded from the voting process. If all of the frames in a submap fail the loop closure detection, this submap is regarded as a undefined submap, which could be a new place that has never been visited or a place with many dynamics. We build the map with the data of Day 1-1 and test the inter-session  loop closure detection method on session Day 3-1 and Day 3-2. The results are shown in Table \ref{loop closure detection}. 

\begin{table}[h]
\caption{The results of the inter-session loop closure detection}
\begin{center}
\label{loop closure detection}
\begin{tabular*}{0.5\textwidth}{p{0.95cm}cccc}
\hline
Sequence & total  & correctly matched & undefined & wrongly matched\\
 & submaps &   submaps&  submaps &  submaps\\ 
\hline
Day 3-1 & 73 & 68 & 5 & 0\\
Day 3-2 & 73 & 73 & 0 & 0\\
\hline
\end{tabular*}\\
\end{center}
\end{table}

\begin{figure}[h]
\centering
\includegraphics[width=0.5\textwidth]{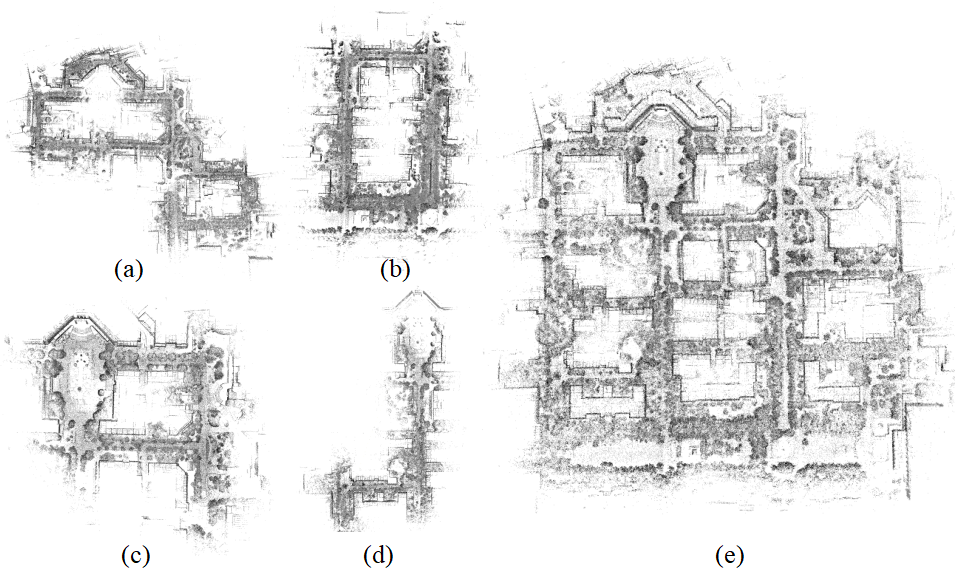}
\caption{The merging result from multi-session maps. (a), (b), (c), (d) are four of the single-session of maps used for map maintenance. (e) is the final merged laser map demonstrating the south of Yuquan Campus. }
\label{multi-session map}
\end{figure}

As the results show, in most of the cases our proposed inter-session loop closure detection method could correctly detect the loop closure. And in these two sequences there is no wrong detection result. We show the environment where the undefined submaps collected from in Fig. \ref{near dynamic}. Since two of the undefined submaps are neighbors, we only show four places in Fig. \ref{near dynamic}. As we can see, in all of the four places, the environment differs due to the parking cars; and these cars are close to the robot when it collects the data. Those close dynamic objects would cause largely changes in the raw laser scans， which leads to the failure of the loop closure detection method.

\subsection{Low dynamics detection in sparse point clouds}

Based on the results of inter-session loop closure detection, we can test the low dynamic detection method on any two of the matched submaps. The length of the voxel grid is set as 0.04 m. We show one of the results in Fig. \ref{low dynamic detection}. As the result shows, the parking cars could be distinguished out as the dynamic objects.  So as some points above the ground which are part of the high dynamic points that are not completely removed by the dynamic detection method in single-session  map construction. But many points on the trees would also be detected. We think this is mainly due to the sparsity of the submaps. While constructing the map from one single-session of data, in order to reduce the computation time and storage, we would down-sample the laser points so the submaps could be sparse. So the point clouds of the same tree but constructed from different sessions would be different, which leads to the response of the dynamic detection method. Since the points on the trees take an important part in laser-based localization method, especially for the outdoor environment, they should not be excluded out of the submaps. So when doing dynamic detection, we set a height threshold of 2m to avoid those points above that being deleted from the submap.

\subsection{Multi-session map construction}

We finally construct a large laser map of the south part of Yuquan Campus, Zhejiang University which covers around 250000 square meter, as shown in Fig .\ref{multi-session map} (e). And Fig .\ref{multi-session map} (a), (b), (c), (d) are four of single-session of maps that used to construct the whole map. For sake of the robustness
of inter-session loop closure detection, each new single session would overlap a large part of the environment with the history map.

\section{CONCLUSIONS}
In this paper we propose a multi-session map construction method for large-scale outdoor environment. The different sessions of data are aligned with a laser-based loop closure detection method. And the redundant information is discarded after the dynamic detection method has been applied between overlapped submaps. The experimental results validate the capability of both the inter-session loop closure detection method and the dynamic detection method.

However, there still exists some problems for those two methods. As for inter-session loop closure detection method, the detection results would be largely influenced by the emergency of new objects which do not exist when the previous data was collected, especially if these object are close to the robot. We want to overcome this problem in the future work by merging some priori information to distinguish the dynamics.

Besides, the problem of the dynamic detection method is that there usually exist some false positive detection results. This problem is partially due to the sparsity of the submap, and partially because this method is sensitive to the precision of the submap alignment. If there is some error in  the estimation of the relative transformation between the overlapped submaps, especially if the error exists in the pitch or roll direction, wrong  detection results would occur. We will try to solve this problem in the future work.

\addtolength{\textheight}{-12cm}   




\section*{ACKNOWLEDGMENT}
We would like to thank the author of \cite{Pomerleau12comp} for releasing the codes of the modular library libpointmathcer\footnote{https://github.com/ethz-asl/libpointmatcher}.


\bibliographystyle{IEEEtran}
\bibliography{library}

\begin{thebibliography}{10}
\providecommand{\url}[1]{#1}
\csname url@rmstyle\endcsname
\providecommand{\newblock}{\relax}
\providecommand{\bibinfo}[2]{#2}
\providecommand\BIBentrySTDinterwordspacing{\spaceskip=0pt\relax}
\providecommand\BIBentryALTinterwordstretchfactor{4}
\providecommand\BIBentryALTinterwordspacing{\spaceskip=\fontdimen2\font plus
\BIBentryALTinterwordstretchfactor\fontdimen3\font minus
  \fontdimen4\font\relax}
\providecommand\BIBforeignlanguage[2]{{%
\expandafter\ifx\csname l@#1\endcsname\relax
\typeout{** WARNING: IEEEtran.bst: No hyphenation pattern has been}%
\typeout{** loaded for the language `#1'. Using the pattern for}%
\typeout{** the default language instead.}%
\else
\language=\csname l@#1\endcsname
\fi
#2}}

\bibitem{pomerleau2014long}
F.~Pomerleau, P.~Kr{\"u}si, F.~Colas, P.~Furgale, and R.~Siegwart, ``Long-term
  3d map maintenance in dynamic environments,'' in \emph{Robotics and
  Automation (ICRA), 2014 IEEE International Conference on}.\hskip 1em plus
  0.5em minus 0.4em\relax IEEE, 2014, pp. 3712--3719.

\bibitem{mcdonald20116}
J.~McDonald, M.~Kaess, C.~Cadena, J.~Neira, and J.~J. Leonard, ``6-dof
  multi-session visual slam using anchor nodes,'' 2011.

\bibitem{schneider2017maplab}
T.~Schneider, M.~Dymczyk, M.~Fehr, K.~Egger, S.~Lynen, I.~Gilitschenski, and
  R.~Siegwart, ``maplab: An open framework for research in visual-inertial
  mapping and localization,'' \emph{arXiv preprint arXiv:1711.10250}, 2017.

\bibitem{yang2013go}
J.~Yang, H.~Li, and Y.~Jia, ``Go-icp: Solving 3d registration efficiently and
  globally optimally,'' in \emph{Proceedings of the IEEE International
  Conference on Computer Vision}, 2013, pp. 1457--1464.

\bibitem{besl1992method}
P.~J. Besl, N.~D. McKay, \emph{et~al.}, ``A method for registration of 3-d
  shapes,'' \emph{IEEE Transactions on pattern analysis and machine
  intelligence}, vol.~14, no.~2, pp. 239--256, 1992.

\bibitem{land1960automatic}
A.~H. Land and A.~G. Doig, ``An automatic method of solving discrete
  programming problems,'' \emph{Econometrica: Journal of the Econometric
  Society}, pp. 497--520, 1960.

\bibitem{dube2017segmatch}
R.~Dub{\'e}, D.~Dugas, E.~Stumm, J.~Nieto, R.~Siegwart, and C.~Cadena,
  ``Segmatch: Segment based place recognition in 3d point clouds,'' in
  \emph{Robotics and Automation (ICRA), 2017 IEEE International Conference
  on}.\hskip 1em plus 0.5em minus 0.4em\relax IEEE, 2017, pp. 5266--5272.

\bibitem{yin2017efficient}
H.~Yin, X.~Ding, L.~Tang, Y.~Wang, and R.~Xiong, ``Efficient 3d lidar based
  loop closing using deep neural network,'' in \emph{Robotics and Biomimetics
  (ROBIO), 2017 IEEE International Conference on}.\hskip 1em plus 0.5em minus
  0.4em\relax IEEE, 2017, pp. 481--486.

\bibitem{walcott2012dynamic}
A.~Walcott-Bryant, M.~Kaess, H.~Johannsson, and J.~J. Leonard, ``Dynamic pose
  graph slam: Long-term mapping in low dynamic environments,'' in
  \emph{Intelligent Robots and Systems (IROS), 2012 IEEE/RSJ International
  Conference on}.\hskip 1em plus 0.5em minus 0.4em\relax IEEE, 2012, pp.
  1871--1878.

\bibitem{fehr2017tsdf}
M.~Fehr, F.~Furrer, I.~Dryanovski, J.~Sturm, I.~Gilitschenski, R.~Siegwart, and
  C.~Cadena, ``Tsdf-based change detection for consistent long-term dense
  reconstruction and dynamic object discovery,'' in \emph{Robotics and
  Automation (ICRA), 2017 IEEE International Conference on}.\hskip 1em plus
  0.5em minus 0.4em\relax IEEE, 2017, pp. 5237--5244.

\bibitem{Pomerleau12comp}
F.~Pomerleau, F.~Colas, R.~Siegwart, and S.~Magnenat, ``{Comparing ICP Variants
  on Real-World Data Sets},'' \emph{Autonomous Robots}, vol.~34, no.~3, pp.
  133--148, Feb. 2013.

\bibitem{lu1997globally}
F.~Lu and E.~Milios, ``Globally consistent range scan alignment for environment
  mapping,'' \emph{Autonomous robots}, vol.~4, no.~4, pp. 333--349, 1997.

\bibitem{K2011G2o}
R.~Kümmerle, G.~Grisetti, H.~Strasdat, K.~Konolige, and W.~Burgard, ``G2o: A
  general framework for graph optimization,'' in \emph{IEEE International
  Conference on Robotics and Automation}, 2011, pp. 3607--3613.

\bibitem{KahlerPM16}
O.~K{\"{a}}hler, V.~A. Prisacariu, and D.~W. Murray, ``Real-time large-scale
  dense 3d reconstruction with loop closure,'' in \emph{ECCV 2016}, 2016, pp.
  500--516.

\end{thebibliography}

\end{document}